\documentclass[sigconf]{acmart}

\renewcommand\footnotetextcopyrightpermission[1]{} 
\usepackage{multirow}
\usepackage{amsmath}
\usepackage{bbm}
\AtBeginDocument{%
  \providecommand\BibTeX{{%
    \normalfont B\kern-0.5em{\scshape i\kern-0.25em b}\kern-0.8em\TeX}}}

\setcopyright{acmcopyright}
\copyrightyear{2018}
\acmYear{2018}
\acmDOI{XXXXXXX.XXXXXXX}

\acmConference[Conference acronym 'XX]{Make sure to enter the correct
  conference title from your rights confirmation emai}{June 03--05,
  2018}{Woodstock, NY}
\acmPrice{15.00}
\acmISBN{978-1-4503-XXXX-X/18/06}

\begin{document}

\title{Adaptive Learning on User Segmentation: Universal to Specific Representation via Bipartite Neural Interaction*\\
{\footnotesize \textsuperscript{*}Note: This work has been published in the proceedings of the 2nd International ACM SIGIR Conference on Information Retrieval in the Asia Pacific.}}


\author{Xiaoyu Tan}
\affiliation{%
 \institution{INF Technology (Shanghai) Co., Ltd}
 \city{Shanghai}
 \country{China}}
 \authornote{Equally Contribution}
\author{Yongxin Deng}
\affiliation{%
  \institution{Shanghai University of Engineering Science}
  \city{Shanghai}
  \country{China}}
  \authornotemark[1]

\author{Chao Qu}
\affiliation{%
  \institution{INF Technology (Shanghai) Co., Ltd}
 \city{Shanghai}
 \country{China}}
 \authornotemark[1]

\author{Siqiao Xue}
\author{Xiaoming Shi}
\author{James Zhang}
\affiliation{%
  \institution{Ant Group}
  \city{Shanghai}
  \country{China}}

\author{Xihe Qiu}
\email{qiuxihe1993@gmail.com}
\affiliation{%
  \institution{Shanghai University of Engineering Science}
  \city{Shanghai}
  \country{China}}
  \authornote{Corresponding Author}




\begin{abstract}
Recently, models for user representation learning have been widely applied in click-through-rate (CTR) and conversion-rate (CVR) prediction. Usually, the model learns a universal user representation as the input for subsequent scenario-specific models. However, in numerous industrial applications (e.g., recommendation and marketing), the business always operates such applications as various online activities among different user segmentation. These segmentation are always created by domain experts. Due to the difference in user distribution (i.e., user segmentation) and business objectives in subsequent tasks, learning solely on universal representation may lead to detrimental effects on both model performance and robustness. In this paper, we propose a novel learning framework that can first learn general universal user representation through information bottleneck. Then, merge and learn a segmentation-specific or a task-specific representation through neural interaction. We design the interactive learning process by leveraging a bipartite graph architecture to model the representation learning and merging between contextual clusters and each user segmentation. Our proposed method is evaluated in two open-source benchmarks, two offline business datasets, and deployed on two online marketing applications to predict users' CVR. The results demonstrate that our method can achieve superior performance and surpass the baseline methods. 
\end{abstract}



\received{20 February 2007}
\received[revised]{12 March 2009}
\received[accepted]{5 June 2009}

\maketitle

\section{Introduction}
Probability prediction is a core technique in both recommendation systems and digital marketing systems. In industrial practice, click-through-rate (CTR), conversion-rate (CVR), and other customer behaviors should be accurately estimated to efficiently provide personalized service \cite{ko2022survey,gao2022graph,dhelim2022survey,rahayu2022systematic}. It should be noted that the performance of these predictions is crucial not only for the quality of the system but also fundamentally determines the revenue of the service provider. 

Recently, there has been an increasing interest in leveraging machine learning (ML) based prediction models in practice. By gathering the data from the history log, a training dataset can be constructed with features and labels. Then, the model can be trained under a supervised learning manner to provide the prediction based on the data knowledge \cite{sun2019bert4rec,wang2017deep}. Utilizing ML models in recommendation, advertisement, and other digital marketing systems has achieved astonishing performance improvement and has been recognized as a standard technique in this field \cite{song2019autoint,guo2017deepfm}.

The growth of digital business has led to an increase in the number of customers and business types, making it increasingly difficult to maintain effective machine learning models. This challenge is even more challenging when the model needs to handle multiple, similar but different subsequent tasks within a large user space. One of the primary difficulties encountered is the diversity of user characteristics, which often results in multi-modal and long-tailed data distribution \cite{liu2020long,oestreicher2012recommendation}. Learning under such data distribution can significantly reduce performance and robustness of machine learning models, particularly when dealing with large and diverse user groups. Another issue is the distributional shift between training data and online inference data. Due to the business expansion, the user base is consistently expanding and changing. Training on historical data and inference on current scenarios is lack timeliness and may induce performance degradation \cite{hendrycks2016baseline,wang2022causal}. 

To ensure service quality under the aforementioned circumstance, business operators start to divide the users into different user segments under business logic and try to build an individual model on each segmentation. For example, operators may mainly segment the user into different user groups by daily activities on the platform and construct individual models on each user segmentation with different marketing settings \cite{wu2009probabilistic,mohapatra2013commerce,zhang2021user}. Although this method can ensure stable performance on each user segmentation or each sub-tasks, it requires an intense workload and a long period to launch in real implementation. When the number of customers is consistently expanding and can be subdivided into more user groups, it is \textit{impractical} to build one model for each group. Another main issue of this method is that each model cannot utilize the meta knowledge learned from other user groups. Limited data for each user group may provide a narrow perspective of data information and reduce the generalization of the model. 

The key aspect of model performance under industry circumstances is representation. If we construct one model on all users, the model would utilize universal representation on all user segmentation and subsequent tasks. If we build one model on each user segmentation or task respectively, the model would leverage the segmentation-specific representation. The former representation contains meta-knowledge learned from all subsequent tasks but may ignore domain-specific details. The latter representation implements domain-specific details on limited data but abandons the meta-knowledge. From this perspective, the user segmentation created by business experts is actually \textit{exploiting the trade-off} between universal and domain-specific representation with prior knowledge to guarantee service performance. \textit{So, can we reduce such exhausted exploitation to an end-to-end learning manner and further improve the performance using ML?}

In this paper, we proposed a novel learning framework that can learn both \textbf{u}niversal and \textbf{g}egmentation-\textbf{s}pecific \textbf{r}epresentation (USSR) in prediction tasks on all data. The USSR learning framework first learns a universal representation in a mixture of Gaussian latent space through information bottleneck. This representation is learned by first projecting user features into different clusters and then representing them in multi-variant Gaussian distribution. This representation can successively encounter the multi-modal and long-tail data distribution and encode the meta-knowledge by learning from all data \cite{rao2019continual}. Then, the model would learn segmentation-specific representation through neural interaction. Since each user segmentation is independent, defined by business operators, and includes domain-specific information, we design a bipartite neural interaction process to better incorporate prior knowledge from business logic and generate segmentation-specific representation from contextual clusters.  Therefore, our proposed method cannot only learn a universal representation from a large dataset, but also adapt new user segmentation for subsequent tasks \cite{huang2021signed,kipf2018neural}. We test our models in two Alipay CVR prediction tasks and deploy them as online services in two Alipay CVR prediction tasks for recommendation and marketing. The online results demonstrate that our proposed method can improve performance compared with single baseline models. We summarize our contribution as follows:
\begin{itemize}
    \item We propose a novel learning framework to learn universal representation and segmentation-specific representation in an end-to-end manner.
    \item We test our framework on two open-source benchmarks and two CVR prediction offline datasets in Alipay. The evaluation demonstrates a significant improvement in predictive accuracy when compared to prior approaches.
    \item We deploy the model as two online CVR prediction services in Alipay and demonstrate high effectiveness and efficiency in practice by observing the online CVR performance. 
\end{itemize}

\section{Related work}
\label{sec:related}
In the realm of recommendation systems and digital marketing, a multitude of approaches have been explored and applied across diverse domains, encompassing the web\cite{castellano2011newer,qu-2022-rltpp,jiang2022learning}, books\cite{porcel2009multi}, tourism\cite{garcia2011sem}, movies\cite{bobadilla2010new}, music\cite{yoshii2008efficient}, resource utilization\cite{xue_meta_2022}, and more. The importance of constructing high-quality proprietary recommendation systems tailored to offer personalized suggestions across different application scenarios cannot be overstated. At the core of recommendation systems and digital marketing lie probability prediction techniques, with extensive research efforts dedicated to precisely estimating CTR, CVR, and other customer behaviors\cite{YANG2022102853,xue-2022-hypro}.

Notably, \cite{guo2017deepfm} introduced DeepFM, an end-to-end learning model designed for CTR prediction. It deftly combines the recommendation prowess of Factorization Machines with the feature learning capabilities of deep learning, addressing the biases towards low-order and high-order features inherent in traditional algorithms. Similarly, \cite{huang2019fibinet} presented FiBiNET, which dynamically models feature importance and effectively learns feature interactions through bilinear functions, achieving outstanding performance in CTR prediction tasks.

However, as user numbers and business diversity grow, data collection often introduces significant noise, leading to the degradation of prediction model performance. Furthermore, models trained on historical data may struggle to adapt to new scenarios, hampering their generalization abilities. Several prior works have made inroads into addressing these challenges.

In contrast to previous approaches, \cite{li2023tdr} introduces a novel targeted double robust method named TDR to mitigate bias and variance issues stemming from error-filled models in existing recommendation systems. \cite{zhang2021user} presents URIPW, a method that confronts the challenge of noisy data due to implicit impression-revisit effects and selection bias. It achieves user retention modeling through a causal perspective by estimating revisit rates. Meanwhile, \cite{su2021attention} introduces a pioneering deep learning framework to tackle complexity and non-linearity issues arising from sparse user history and significant delays between click and conversion behaviors.

Moreover, \cite{kolesnikov2012predicting} diverges from the norm by employing events, such as clicks and skips, rather than observed CTR, to predict ads when historical click records are lacking. \cite{kumar2015predicting} establishes a model for predicting advertisement CTRs using Logistic Regression, effectively representing and constructing conditions and vulnerabilities among variables. These prior research endeavors predominantly concentrate on mitigating data noise, addressing multimodality, managing long-tail distributions, and handling distribution shifts.

Furthermore, some works divide users into different user segments and improve the performance of models in large-scale industrial recommendation scenarios by constructing individual models suitable for different user segments. For example, \cite{boratto2016using} analyzes user ratings and extracts word embedding representations to build reliable user segments.

What distinguishes our work from this comprehensive backdrop is the introduction of a novel learning framework, USSR. This framework facilitates end-to-end learning of both general representations and fine-grained user-specific representations. Unlike prior efforts, which often necessitate a trade-off between the two, USSR initially leverages information bottleneck learning to extract general representations that encapsulate meta-knowledge derived from the entire dataset. Previous methods typically confined representation learning to individual datasets.

USSR further distinguishes itself by employing a dual-part neural interaction mechanism to acquire user-specific representations, allowing for more effective integration of domain expert prior knowledge—an integration that has historically proven challenging. This enables our model to adapt dynamically to the evolving data landscape, a feat not easily accomplished by traditional representation learning methods.

Our framework has been rigorously evaluated across multiple real-world business scenarios and has been successfully deployed in Alipay's online services. In contrast, previous representation learning research primarily remained confined to theoretical exploration and benchmark datasets.

Notably, USSR significantly enhances prediction performance, surpassing approaches that exclusively focus on learning either general or user-specific representations. This underscores the advantages of simultaneously learning both types of representations, solidifying the uniqueness of our approach.

\section{Methods}\label{sec:methods}
\begin{figure*}[h] 
    \centering \includegraphics[width=1\textwidth]{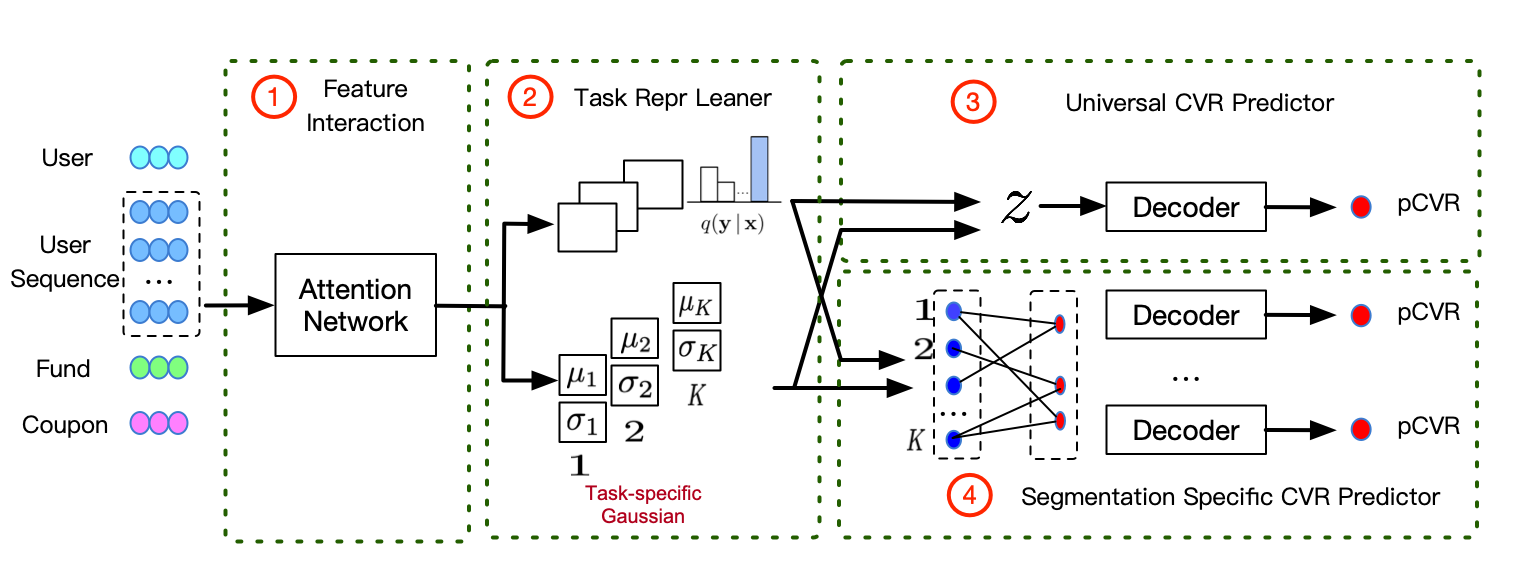}
    \caption{The architecture of USSR framework used in CVR prediction tasks.} 
    \label{fig:ussr}
\end{figure*}

The USSR learning framework first learns the universal representation of all data through information bottleneck on a mixture of Gaussian latent space. Then, the segmentation-specific representation would be learned by bipartite neural interaction between contextual representation and domain-specific representation. The architecture of the USSR framework used in CVR prediction tasks is shown in Figure \ref{fig:ussr}.

\subsection{Preliminary}
For probability prediction in digital marketing tasks (e.g., CTR and CVR tasks), the model always observes a feature vector $x \in \mathbbm{R}^n$ with $n$ dimension in dataset $\mathcal{D}$. In practice, $x$ is always concatenated by user features, item features, and features from other domains. The model can be optimized by using cross-entropy loss on the training dataset between true labels $y$ and model prediction $\hat{y} = f(x)$:
\begin{equation}
    \mathcal{L} = \mathbbm{E}\left[y{\rm log}\hat{y}+(1-y){\rm log}(1-
    \hat{y})\right].
\end{equation} 
The representation $z$ for each data $x$ can be learned either implicitly or explicitly \cite{kingma2013auto,kipf2016variational,pu2016variational,vahdat2020nvae} by using numerous feature extraction or feature interaction techniques \cite{guo2017deepfm,wang2017deep,song2019autoint,sun2019bert4rec}. To perform marketing operations, the operators may maintain multiple models $\{f_i: i \in (1...M)\}$ on $M$ different user segmentation with domain-specific feature $d_i$. The prediction result for each segmentation is separately predicted by the model $\hat{y}_i=f_i(x_i \mathbin\Vert d_i)$, where $\mathbin\Vert$ denotes the concatenation operation. These segmentation can indeed improve the overall performance but unfortunately, ignore the meta- knowledge between different segmentation.

\subsection{Universal Representation}
\label{sec:universal}
The universal representation can be learned on all data with information bottleneck. This learning process would compress the information from all features and generate task-relevant representation \cite{rao2019continual}. The objective function of this learning process is:
\begin{equation}
\label{equ:1}
    \mathcal{L} = I(z,y)-\beta I(z,x),
\end{equation}
where $\beta$ is Lagrange multiplier and $I$ represents the mutual information. By maximizing the objective function, the task-relevant information can be reserved and irrelevant information can be discarded. To capture the multi-modal distribution and potential distributional shift due to continual operation, we encode the user feature in a mixture of Gaussian latent space. We use $q(c|x)$ to map $x$ into $c$ clusters where $c$ is categorical variables. Then, the representation $z$ can be sampled from the Gaussian distribution with the parameter generated by model $q(z|x, c=k)$ on component $k$ with a single multi-variant Gaussian encoder head. Each head has a specific prior normal distribution with parameters $\mu_z (c)$ and $\sigma_z (c)$, which represent the mean and variance, respectively. Therefore, the encoding process is to choose one Gaussian encoding head by learning a multi-layer perceptron with softmax layer output and represent the embedding using head-specific parameter $z\sim \mathcal{N}(\mu_z (c=k), \sigma_z (c=k))$. After acquiring  $z$, we can use one decoder $p(y|z)$ to perform universal target prediction. 

However, the objective function shown in Eqn~\eqref{equ:1} is intractable and cannot be directly optimized. Instead, we corporate the aforementioned model structure and optimize the evidence lower bound by minimizing the following objective function:
\begin{equation}
\label{equ:lower}
\begin{aligned}
    \mathcal{L}_{\rm ib} = & \sum_{k=1}^K q(c=k|x) [ {\rm log}p(y|\tilde{z}_k) - \\  & {\rm KL}(q(z|x, c=k)||p(z|c=k))] - {\rm KL}(q(c|x)||p(c)),
\end{aligned}
\end{equation}
where $\tilde{z}_k \sim q(z|x, c=k)$ and ${\rm KL}$ represents the KL-divergence between two distribution. To achieve an accurate and informative representation of complicated feature structures, variant feature interaction models can be implied as encoder and decoder of our framework. In our implementation, we utilize AutoInt \cite{song2019autoint} as the encoder. 


\subsection{Segmentation-specific Representation}
\label{sec:seg}
Leveraging the universal representation acquired in Section \ref{sec:universal} on all subsequent tasks is feasible but not accurate. In practical digital marketing scenarios, the operators always perform domain-specific marketing under different user segmentation and believe that such segmentation is marketing navigation to improve the total business performance (e.g., constructing different coupon sets based on age, gender, or phone systems). Incorporating prior knowledge into machine learning models has been shown to improve system performance. However, this often requires domain-specific modeling on different user distributions, which can be time-consuming \cite{lee2019melu, lu2020meta}. To better incorporate the prior knowledge from business experts and leverage the informative universal representation on domain-specific tasks, we design an adaptive neural interaction learning process similar to \citet{kipf2018neural} to perform representation interaction between universal representation and segmentation-specific knowledge. Since the user-segmentation defined by the operators can contain one or more contextual groups $c$ learned from Section \ref{sec:universal}, we construct the neural interaction process under a bipartite graph architecture \cite{huang2021signed}. 



Suppose the whole user group is divided into $M$ user-segmentation with segmentation-specific features:  $\{u_m: m \in (1...M)\}$ and the users can be represented into $K$ contextual groups with latent representation : $\{z_k: k \in (1...K)\}$. The $u_m$ can be defined at the domain level or user level and contains prior knowledge from experts. The neural interaction process under bipartite architecture requires edge encoder $f_e$, vortex encoder $f_v$, edge decoder $\hat{f}_e$, and vortex decoder $\hat{f}_v$. 

We first use an edge encoder $f_e$ to learn the first interactive representation of all possible interactions:
\begin{equation}
    h_{k,m}^{1} = f_e(z_k \mathbin\Vert u_m).
\end{equation} 

Then, the information can be summarized from both contextual and segmentation perspectives through vortex encoders: 
\begin{equation}
\begin{aligned}
    & h_{k}^{2,M} = f_v(\sum_{m=1}^M h_{k,m}^{1}),
    & h_{m}^{2,K} = f_v(\sum_{k=1}^K h_{k,m}^{1}).
\end{aligned}
\end{equation} Finally, the bipartite interaction architecture can be learned by edge encoder again and sparsely represented in $e_{k,m}$ through a softmax layer. By implementing the bipartite architecture, the segmentation-specific representation $\hat{h}_m$ can be learned by edge decoder:
\begin{equation}
\hat{h}_m = \sum_{k=1}^{K}e_{k,m}\hat{f}_e(z_k \mathbin\Vert u_m),
    e_{k,m} \sim {\rm softmax}(f_e(h_{k}^{2,M}\mathbin\Vert h_{m}^{2,K})).
\end{equation} 


Based on the implementation details on subsequent tasks, the prediction can be made by one vortex decoder or $M$ decoders for each user-segmentation: $\{\hat{f}_{v,m}: m \in (1...M)\}$. We recommend using multiple vortex decoders to achieve robust prediction results during the segmentation expansion: $\hat{y} = \hat{f}_{v,m}(\hat{h}_m).$ This architecture can be typically optimized following the cross-entropy loss between $\hat{y}$ and true label $y$ introduced in equation \ref{equ:1}.  

\subsection{Adaptive Learning on Segmentation and Data Expansion}
Real industry applications always require the models to have the elasticity to adapt data expansion \cite{chong2022literature}. Business operators also always create and update user segmentation based on their business understanding and marketing processes. The USSR learning framework can increase the capacity of universal representation by dynamic expansion on contextual clusters $c$. The segmentation-specific representation can also be incrementally trained without influencing the model parameters. 


For the dynamic expansion of universal representation, we set threshold $t_{\rm logit}$ on the log-likelihood evaluation with equation \ref{equ:lower} and threshold $t_{\rm num}$ on data numbers. If the evaluation result of new data is larger than $t_{\rm logit}$, we put the data into a new data buffer $\mathcal{D}_{\rm new}$. If the data number of $\mathcal{D}_{\rm new}$ is larger than $t_{\rm num}$, we perform few shot iterative training through loss function:
\begin{equation}
\label{equ:expansion}
\begin{aligned}
    \mathcal{L}_{\rm ed} = & {\rm log} q(c=K+1|x) + {\rm log} p(y|\tilde{z}_{K+1}) \\ & - {\rm KL}(q(z|x, c=K+1)||p(z|c=K+1))].
\end{aligned}
\end{equation} The component $q(z|x, c=K+1)$ can be initialized by $q(z|x, c=l^*)$ where:
\begin{equation}
    l^* = \mathop{{\rm argmax}}\limits_{l \in \{1,2,...,K\}}\sum_{x\in \mathcal{D}_{\rm new}}q(c=l|x).
\end{equation}

For the expansion of segmentation-specific representation, the whole process is controlled by business operators and the learning process can be disentangled from other representations. This can be achieved by  first storing the segmentation-specific representation $\{\hat{h}_m : m\in(1,...,M)\}$ in $\mathcal{D}$ and solely performing inference by $\hat{f}_{v,m}$. This process can also significantly reduce the model inference duration. Then, we can repeat the whole learning process of Section \ref{sec:seg} to get a new segmentation-specific representation $\hat{h}_{M+1}$ and new inference model $\hat{f}_{v, M+1}$. This process can ensure the new representation contains the meta-knowledge from all previous user segmentation but all previous representation remains the same. Moreover, the robustness and stability of online performance can also be satisfied during the model industrial launch.

\section{Experiments and Performance}
\label{sec:exp}

Our experiment aims to investigate the effectiveness of the proposed framework in benchmarks, real-world scenarios, and its potential for practical applications. The experiment can be divided into three main parts. Firstly, we evaluate the performance of our USSR framework on two widely used public datasets, Criteo\cite{Criteo} and Avazu\cite{Avazu}, and compare it with state-of-the-art algorithms. Secondly, we assess the performance of the USSR framework on two distinct offline datasets derived from Alipay's industrial application. Lastly, to evaluate the framework's ability to adapt to real-time data (which we consider the most significant capability), we deploy the framework online on two Alipay services. The latter two parts of the experiment focus on different user group segmentation CVR prediction tasks in the Alipay Caifu and Jiebei scenarios. These scenarios face challenges such as lone-tile data distribution, expanding user space, and increasing user segmentation (e.g., one scenario contains more than 20 user segmentation). For these scenarios, due to the model deployment policies of the company and resources limitation, we only choose the model originally used in the online service, a universally trained AutoInt model \cite{song2019autoint}, as our baseline method. For public evaluation, we choose multiple state-of-the-art methods as our baseline methods to fully evaluate our proposed algorithm.

\subsection{Evaluation on Public Datasets}
The Criteo dataset is a widely used public dataset for research and development in advertising recommendation systems and machine learning algorithms. It is provided by Criteo and contains anonymized online advertising transaction data, where each record consists of a series of features and labels. The features include anonymized user identifiers, properties of the ads (e.g., category, size, format), historical CTR of ads, contextual information of ad placements, etc. Specifically, it includes 13 dense features and 26 sparse features, and the label indicates whether the user clicked on the ad. The Avazu dataset, on the other hand, is a public dataset of online ad click data commonly used for CTR prediction and advertising recommendation machine learning algorithms. It contains billions of ad transaction records, where each record includes various features and labels. The features include user and ad information, such as anonymized user identifiers, ad IDs, ad categories, time, and geographic location of ad placements, primarily consisting of sparse features. The label represents whether the user clicked on the ad, used for building binary classification machine learning models.

\begin{table}
  \caption{The experiment results of AUC on two public CTR tasks datasets compared with baseline methods.}
  \label{tab:public}
\begin{tabular}{c|c|c}
\hline
\textbf{Data} & \textbf{Methods} & \textbf{AUC}    \\ \hline
Avazu         & DeepFM\cite{guo2017deepfm}           & 0.7836          \\ \cline{2-3} 
              & xDeepFM\cite{lian2018xdeepfm}          & 0.7855          \\ \cline{2-3} 
              & DCN\cite{wang2017deep}              & 0.7681          \\ \cline{2-3} 
              & ONN\cite{yang2020operation}              & 0.7513          \\ \cline{2-3} 
              & FiGNN\cite{li2019fi}            & 0.8120          \\ \cline{2-3} 
              & AutoInt\cite{song2019autoint}          & 0.7752          \\ \cline{2-3} 
              & USSR(ours)       & \textbf{0.8133} \\ \hline
Criteo        & DeepFM           & 0.8085          \\ \cline{2-3} 
              & xDeepFM          & 0.8091          \\ \cline{2-3} 
              & DCN              & 0.8067          \\ \cline{2-3} 
              & ONN              & \textbf{0.8123} \\ \cline{2-3} 
              & FiGNN            & 0.8082          \\ \cline{2-3} 
              & AutoInt          & 0.8061          \\ \cline{2-3} 
              & USSR(ours)       & 0.8068          \\ \hline
\end{tabular}
\end{table}

For our experiments, we used the datasets provided by PaddleRec\cite{bi2022paddlepaddle,paddlerec}. For the dense features in the dataset, we applied a logarithm transformation to restrict their range. As for the sparse features, we first calculated the frequency of each feature category occurring in the entire dataset. Then, we assigned integer indices to the categories based on their ranking in terms of frequency (the indices in the test set correspond to those in the training set, so in the actual processing, we concatenate the training and test sets firstly, generate the indices, and then split them). Since the features in the dataset exhibit a long-tail distribution, we set an upper limit for the indices. Finally, we used these indices to generate embedding vectors for training. In this experiment, we conduct the comparative analysis between our proposed USSR and several state-of-the-art approaches that have been extensively employed in practical applications. Specifically, we consider DeepFM \cite{guo2017deepfm}, xDeepFM \cite{lian2018xdeepfm}, DCN\cite{wang2017deep}, ONN\cite{yang2020operation} , FiGNN \cite{li2019fi} , and AutoInt \cite{song2019autoint} as the benchmark methods for our evaluation. 

The experimental results in terms of Area Under the Curve (AUC) are presented in Table \ref{tab:public}, along with the performance of some state-of-the-art models for comparison. Please note that for the evaluation on public datasets, we only tested the universal representation learning and segmentation-specific representation learning parts introduced in Section \ref{sec:universal} and Section \ref{sec:seg}. The results of our study demonstrate that the USSR framework exhibits superior performance compared to other baseline models when employed on the Avazu dataset, yielding an AUC of 0.8133. This value significantly surpasses the majority of baseline methods, which generally yield AUC scores below 0.80. In the Criteo dataset experiment, our proposed method also demonstrates remarkable performance, surpassing an AUC score of around 0.81 and closely rivaling other state-of-the-art baseline methods, positioning it among the top-performing practical solutions available.

\subsection{Evaluation on Alipay Offline Datasets}
For the offline tests, we extracted the data from historical log records, including user features, user segmentation index, and CVR labels, over a span of 30 days. We randomly shuffled and split the dataset into a training set and a validation set, using 80\% and 20\% of the first 27 days of data, respectively. The remaining 3 days' data was kept as the test set. We trained the proposed model on the training set and evaluated its performance on the validation set. The test set remained unseen until the final evaluation. To prevent overfitting, we employed early stopping based on the performance on the validation set. Similar to the evaluation on public datasets, for the offline evaluation, we only tested the universal representation learning and segmentation-specific representation learning parts introduced in Section \ref{sec:universal} and Section \ref{sec:seg}. The experimental results in terms of AUC are described in Table \ref{tab:offline}. Additionally, we reported the worst performance of the models among all available segmentations, which indicates the robustness of segmentation-specific training. Our results demonstrate that the USSR framework outperforms previous approaches in both overall and segmentation-specific performance measures in the offline tests.

\begin{table}
\caption{The experiment results of AUC on two offline CVR tasks datasets compared with baseline method.}
\label{tab:offline}
\begin{tabular}{c|c|c|c}
\hline
\textbf{Data}                & \textbf{Methods} & \textbf{AUC}  & \textbf{AUC(Worst)} \\ \hline
\multirow{2}{*}{Dataset   1} & AutoInt          & 0.88          & 0.67                \\ \cline{2-4} 
                             & USSR(ours)             & \textbf{0.90} & \textbf{0.76}       \\ \hline
\multirow{2}{*}{Dataset   2} & AutoInt          & 0.80          & 0.70                \\ \cline{2-4} 
                             & USSR(ours)             & \textbf{0.84} & \textbf{0.78}       \\ \hline
\end{tabular}
\end{table}

\subsection{Evaluation on Alipay Online Scenarios}
We deployed the model online and monitored its performance for 7 days by serving 10\% of online customers in two online scenarios. Our trained models only provided CVR prediction results, while item recommendation was handled by the Alipay internal online optimization service, which can set cost constraints \cite{zhou2021antopt}. To ensure a fair comparison, we set the cost constraint of the two models on the same scale. Similar to the offline tests, we compared the performance with the main operating model (i.e., AutoInt model), directly observed the mean CVR results of each service, and examined the robustness by reporting the worst performance of the USSR framework among all segmentations. The results are presented in Table \ref{tab:online}. Based on the online performance, we observed that the trained models under the USSR framework achieved higher online performance and significant improvement even on the worst segmentation. Finally, the trained models passed the online A/B test and were deployed as the main models in these two scenarios.

\begin{table}
\caption{The online results of CVR on two online CVR prediction services compared with baseline method.}
\label{tab:online}
\begin{tabular}{c|c|c|c}
\hline
\textbf{Service}           & \textbf{Methods} & \textbf{CVR}    & \textbf{CVR(Worst)} \\ \hline
\multirow{2}{*}{Service 1} & AutoInt          & 1.88\%          & 1.57\%              \\ \cline{2-4} 
                           & USSR(ours)             & \textbf{2.12\%} & \textbf{1.98\%}     \\ \hline
\multirow{2}{*}{Service 2} & AutoInt          & 1.19\%          & 1.03\%              \\ \cline{2-4} 
                           & USSR(ours)             & \textbf{1.37\%} & \textbf{1.22\%}     \\ \hline
\end{tabular}
\end{table}

\section{Conclusion}
\label{sec:con}
In this paper, we propose a novel learning architecture that can perform universal representation and user-segmentation representation (USSR) learning adaptively in an end-to-end manner. Our proposed approach employs a deep information bottleneck to learn universal representation, while leveraging neural interaction to acquire segmentation-specific representation. Additionally, the framework dynamically expands the representation space to accommodate data expansion. To evaluate the effectiveness of our approach, we first evaluate and compare with other state-of-the-art methods on the open-source dataset. Then, we applied it to two distinct Alipay applications and deployed the models in two Alipay online services. The experiment results and online performance demonstrate that the USSR approach can adapt to different user groups and enhance its performance in an efficient end-to-end manner.
The proposed approach has the potential to improve the performance of various applications in real-world settings. Future research may further explore the capabilities and limitations of the USSR approach and investigate its potential applications in various domains.

\bibliographystyle{ACM-Reference-Format}
\bibliography{sample-base}

\end{document}